%% file: root.tex
\documentclass[letterpaper, 10 pt, conference]{ieeeconf}  

\IEEEoverridecommandlockouts                              

\usepackage[backend=biber,
            hyperref=true,
            url=false,
            isbn=false,
            doi=false,
            backref=false,
            style=ieee,
            citestyle=numeric-comp,
            sorting=none, 
            block=none]{biblatex}

\usepackage{amsmath} 
\usepackage{amssymb}  
\usepackage{multirow}
\usepackage{graphicx}
\usepackage{array}
\usepackage{float}
\usepackage{color}
\usepackage{hyperref}
\usepackage{cleveref}
\usepackage{arydshln}
\usepackage{url}

\newcommand{\wid}{0.14}

\newcommand{\modela}{{\small \texttt{NDDS DOPE}}}
\newcommand{\modelb}{{\small \texttt{NDDS DOPE full}}}
\newcommand{\modelc}{{\small \texttt{NViSII DOPE}}}
\newcommand{\shrink}{-10}

\title{\LARGE \bf
Fast Uncertainty Quantification for Deep Object Pose Estimation
}

\author{Guanya Shi$^{1,2}$, Yifeng Zhu$^{1,3}$, Jonathan Tremblay$^{1}$, \\
Stan Birchfield$^{1}$, Fabio Ramos$^{1,4}$, Animashree Anandkumar$^{1,2}$, Yuke Zhu$^{1,3}$
\thanks{$^{1}$NVIDIA, $^{2}$California Institute of Technology, $^{3}$The University of Texas at Austin, $^{4}$The University of Sydney.}
}

\begin{document}

\maketitle
\thispagestyle{empty}
\pagestyle{empty}

\begin{abstract}

Deep learning-based object pose estimators are often unreliable and overconfident especially when the input image is outside the training domain, for instance, with sim2real transfer. Efficient and robust uncertainty quantification (UQ) in pose estimators is critically needed in many robotic tasks. In this work, we propose a simple, efficient, and plug-and-play UQ method for 6-DoF object pose estimation. We ensemble 2--3 pre-trained models with different neural network architectures and/or training data sources, and compute their average pairwise disagreement against one another to obtain the uncertainty quantification. We propose four disagreement metrics, including a learned metric, and show that the average distance (ADD) is the best learning-free metric and it is only slightly worse than the learned metric, which requires labeled target data. Our method has several advantages compared to the prior art: 1) our method does not require any modification of the training process or the model inputs; and 2) it needs only one forward pass for each model. We evaluate the proposed UQ method on three tasks where our uncertainty quantification yields much stronger correlations with pose estimation errors than the baselines. Moreover, in a real robot grasping task, our method increases the grasping success rate from 35\% to 90\%. Video and code are available at \href{https://sites.google.com/view/fastuq}{https://sites.google.com/view/fastuq}.

\end{abstract}

\input{intro}

\input{related}
\input{method}
\input{experiments}

\input{conclusions}

\printbibliography
\end{document}

%% file: intro.tex
\section{Introduction}
\begin{figure}[t]
  \centering
  \includegraphics[width = 0.9\linewidth]{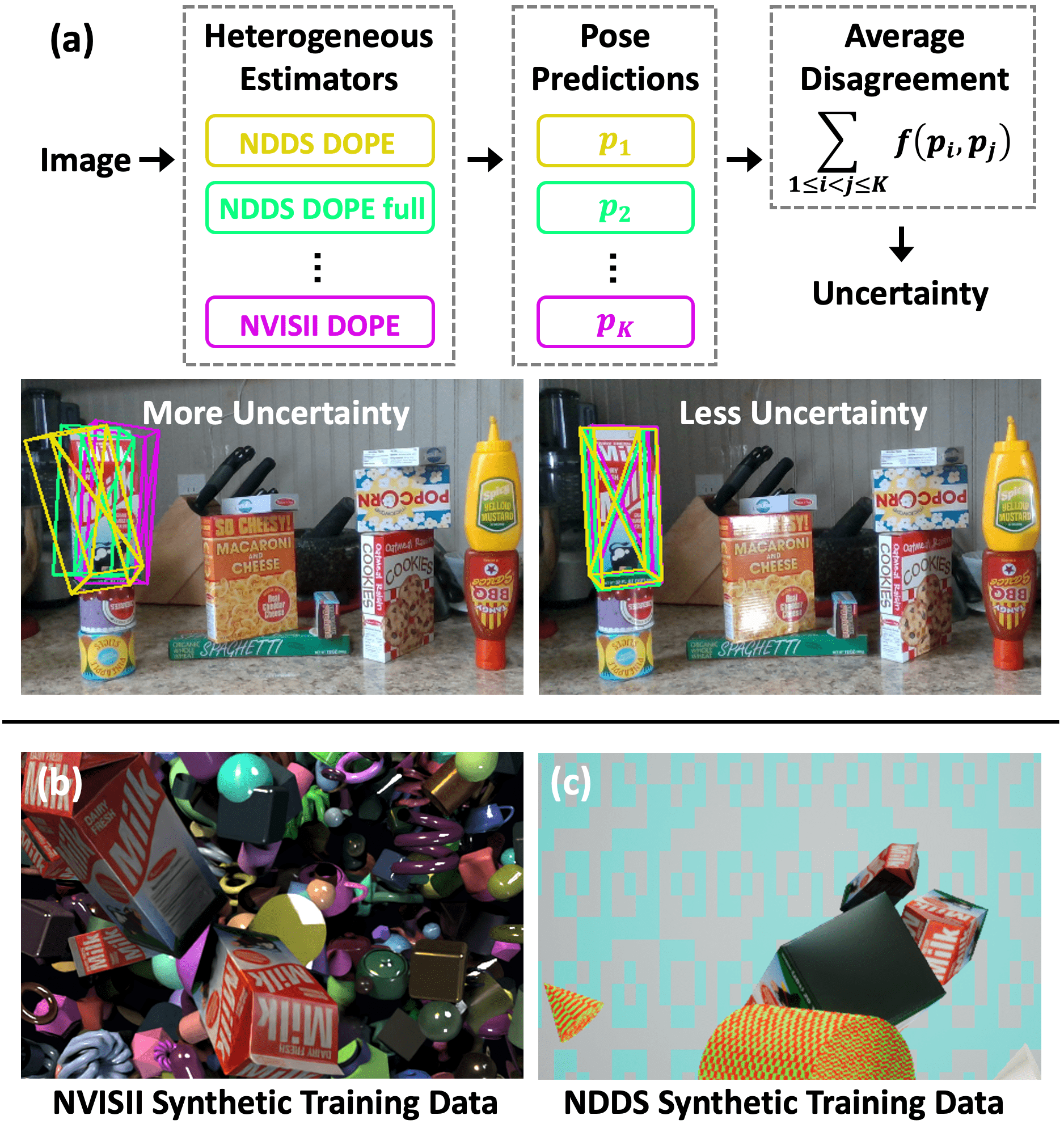}
  \vspace{-8pt}
  \caption{(a) Illustration and examples of using an ensemble of heterogeneous models for uncertainty quantification. We calculate the average disagreement of $K$ pose predictions from $K$ different estimators as an estimation of uncertainty. The examples show the Milk object results of two images in the real-world HOPE dataset~\cite{tyree2019hope}. (b-c) Milk object examples in the synthetic NViSII (b) and NDDS (c) training datasets.}
  \label{fig:ensemble}
  \vspace{\shrink pt}
\end{figure}


Robust and efficient estimation of object poses is a fundamental robot perception task that enables robots to make autonomous decisions in unstructured environments. Recent advances in deep learning-based machine perception have substantially improved the accuracies of pose estimation from real-world sensory data~\cite{tremblay2018pose,xiang2018rss:posecnn,wang2019densefusion,tan20176d,zeng2017multi}. The estimated 6-DoF object pose, represented by the translation and rotation in \emph{SE}(3), serves as a compact and informative state representation for a variety of downstream tasks, such as robot grasping and manipulation~\cite{chen2019grip}, human-robot interactions~\cite{yang2020human}, online camera calibration~\cite{lee2020icra:dream}, and tele-presence robot control~\cite{handa2020dexpilot}.

%


While these deep learning-based pose estimation models have attained remarkable performance, their applicability in mission-critical robotic tasks is hindered by their brittleness in dealing with inputs that are outside the training domain~\cite{amodei2016concrete} or with perturbations in the visual observations~\cite{loquercio2020general}. These issues are especially relevant in pose estimation with sim2real transfer~\cite{doersch2019sim2real,tremblay2018pose,lu2020robust}, where deep-learning models trained on synthetic data tend to be often wrong but overconfident on real data.
To enhance the reliability of deep pose estimation models in risk-sensitive domains, it is critical to endow these models with the capability of accurately assessing the \emph{uncertainty} of their own predictions and identifying failure cases automatically.


Uncertainty quantification (UQ) in robotics has been a long-standing research topic for many decades~\cite{kalmanfilter} due to the strict safety and reliability requirements of real-world systems.
%
%
UQ can take on different meanings in robotic applications. 1) UQ can be in the form of a \emph{probabilistic} distribution of possible outcomes on a prediction. 2) Alternatively, UQ can be a \emph{deterministic} confidence measure of a specific prediction.
In this work, we study the latter case, {\em i.e.}, we focus on a deterministic approximation of the quality of a certain pose prediction (see the formal definition in \Cref{sec:spearman}). We refer to this type of UQ as \emph{predictive uncertainty}, which is easier to estimate and use than the distribution of possible outcomes, since it is a scalar quantity and it directly reflects the risk of a certain pose estimation.

For the first type, Deng {\em et al.} estimated the distributions of possible poses for different objects, such as objects with symmetries, using particle filters~\cite{deng2019poserbpf}. Lee {\em et al.} (our baseline) also estimated the distributions of pose estimations by sampling the belief maps of the neural network output~\cite{lee2020guided}. Such distributions can be embedded with other modules such as exploration in reinforcement learning~\cite{lee2020guided}.  
For the second type, one example is~\cite{tremblay2018pose} (our baseline), where a trained deep object pose estimation model directly outputs a confidence score via keypoint belief maps. However, both the distribution estimation~\cite{lee2020guided} and the confidence estimation~\cite{tremblay2018pose} are not necessarily trustworthy and there is often a mismatch with the true uncertainty, as demonstrated in this paper.


Our goal is to develop a robust and computationally efficient method for quantifying the uncertainty of 6-DoF pose predictions from deep object pose estimation models. This is a challenging task from several directions. First, the predictive uncertainty is difficult to evaluate, since the ground truth for such measurements is non-existent. It thus requires us to develop a practical metric to quantify the uncertainty estimates. Second, deep pose estimation models are typically trained on synthetically rendered data~\cite{doersch2019sim2real,tremblay2018pose}, due to the prohibitive costs of annotating 3D bounding boxes on a large set of real images. This sim2real gap requires the UQ method to robustly handle the discrepancies between the simulated data and real images and to make reliable estimates of predictive uncertainty across the two domains.

\textbf{Contributions.} We develop a simple, efficient, and plug-and-play technique for estimating predictive uncertainty using an ensemble of 2--3 pre-trained models with different architectures and/or training data sources, see \Cref{fig:ensemble}(a). Unlike prior approaches~\cite{loquercio2020general,gal2016dropout,Andrieu2003}, our method does not require any modification of the training process or the model input, and our method only takes one forward pass for each model at inference time. 
Thus, we can readily integrate our UQ method with off-the-shelf deep pose estimation methods~\cite{tremblay2018pose}. Our method uses only a few (2--3) deep pose estimators, computes their average pairwise disagreement to estimate the predictive uncertainty. We propose four metrics for computing the disagreement: translation, rotation, average distance (ADD) and a learned metric. We use the Spearman's rank correlation coefficient to choose the best disagreement metric for predictive uncertainty. This is the first time, to the best of our knowledge, that such a correlation measure is used to quantify the extent to which the predictive uncertainty is consistent with the true uncertainty.


In our experiments, we evaluate the proposed UQ method on three tasks. In the first task, we evaluate the correlation between the estimated predictive uncertainties and pose estimation errors on the HOPE dataset~\cite{tyree2019hope}. Our UQ method has much stronger Spearman's correlations than the baselines. Using the Spearman's correlation, we find that  ADD is the best data-free disagreement metric and that it is only slightly worse than the learned metric, which requires labeled target data. In the second task, we apply the proposed UQ to a camera perspective selection task. We evaluate the performance on a synthetic dataset generated by a photorealistic renderer NViSII~\cite{Morrical20visii}. The most confident camera perspectives selected by our method lead to a 30--40\% reduction of pose estimation errors over the baselines. Finally, we demonstrate that our UQ method trained only on simulated data can guide a real robot to select the best point of view to robustly grasp objects. Our method dramatically improves the grasping success rate from 25\% (by baseline~\cite{tremblay2018pose}) or 35\% (by~\cite{lee2020guided}) to 90\%.





%% file: related.tex
\section{Related Work}

\subsection{Object Pose Estimation} 
The problem of object pose estimation is vibrant within the robotics and computer vision 
communities \cite{tremblay2018pose,hinterstoisser2012model,hodan2017wacv:tless,zakharov2019dpod,xiang2018rss:posecnn,hu2019segmentation,peng2019pvnet,Sundermeyer_2018_ECCV,tekin2018cvpr:objpose}.
Recent leading methods rely on an approach similar to the one used in our work:  A network is trained to predict object keypoints in the 2D image, followed by  
P$n$P \cite{lepetit2009ijcv:epnp} to estimate the pose of the object in the camera coordinate frame \cite{tremblay2018pose,peng2019pvnet,tan20176d,hu2019segmentation,tekin2018cvpr:objpose,dhall2019arx:cones}. 
Other methods have regressed directly to the pose \cite{xiang2018rss:posecnn,Sundermeyer_2018_ECCV}, but these methods bake the camera intrinsics into the learned weights, although geometric post-processing can address this limitation~\cite{Sundermeyer_2018_ECCV}. 
In a related strand, researchers have used keypoint detection for human pose estimation \cite{wei2016cvpr:cpm,cao2017cvpr:mppaf,xiao2018simple,li2019rethinking,sun2019deep}. 
Nevertheless, in robotics applications, it is not uncommon for objects to be detected via fiducial markers~\cite{liu2018:robotsafe,kim2019icra:insertion,tian2019icra:fog}.
Closely related to our work, Deng {\em et al.}~\cite{deng2019poserbpf} used a particle filter-based approach to predicting the distribution of possible object poses.
In contrast, our work studies pose estimators with deterministic predictions, and we quantify and validate their uncertainty.

\subsection{Uncertainty Quantification (UQ)}
Several methods have been proposed to quantify uncertainty in deep learning models. A relatively simple method is to directly use the output confidence of the trained model~\cite{tremblay2018pose}.
However, this has been shown to have limited robustness \cite{blum2019fishyscapes}, and often yielding overconfident estimates in our experiments.
A more principled approach is to adopt Bayesian principles~\cite{bernardo2009bayesian} and posterior inference approximations such as the dropout approximation~\cite{gal2016dropout}, variational inference~\cite{Jordan99}, and Markov chain Monte Carlo~\cite{Andrieu2003}. In general, Bayesian approaches require the specification of prior distributions over model parameters and are more computationally expensive. In contrast, alternative sampling approaches~\cite{loquercio2020general} can estimate uncertainty without changing the network training process by adopting a frequentist strategy. However, it requires multiple forward passes and significantly increases the computational cost during inference. 
Ensemble methods~\cite{Lakshminarayanan:ensemble}, on the other hand, are much easier to run at inference time with modern parallel computational resources. However, they still require modifications in the training procedure and the members in the ensemble need to have the same architecture.  
To the best of our knowledge, ensemble-based UQ has not been previously explored in context of object pose estimation. We leverage this idea but consider a much simpler class of ensembles through disagreement metrics. 


%% file: method.tex
\section{Problem Statement and Method}
We first define the 6-DoF pose estimation problem and the uncertainty quantification (UQ) problem. 
We also introduce the metrics used for UQ. 
Then we introduce our ensemble-based method in \Cref{sec:ensemble,sec:learned}, as depicted in \Cref{fig:ensemble}(a). \Cref{sec:ensemble} formulates our ensemble method with a general disagreement metric, and \Cref{sec:learned} focused on a special case where the metric is learned.   

\subsection{Problem Statement and Metrics}
\label{sec:spearman}
\subsubsection{6-DoF pose estimation and ADD}
In the 6-DoF pose estimation task, the pose estimator $g$ takes an image $x$ as the input, and outputs a deterministic pose prediction $p=g(x)$ in \emph{SE}(3) -- rotation and translation.
We focus on model-based pose estimation of known objects, where each object has a separate trained pose estimator. 
Suppose the ground truth pose is given by $\bar{p}$. 
We measure the error of the pose estimation $p$ by the average distance (ADD)~\cite{hinterstoisser2012model} between poses $p$ and $\bar{p}$,  which is denoted by $\texttt{ADD}(p,\bar{p},\mathcal{X})$:
\[
\texttt{ADD}(p,\bar{p},\mathcal{X}) = \frac{1}{|\mathcal{X}|}\sum_{t \in \mathcal{X}} \left\Vert (p_R t + p_T) - (\bar{p}_R t + \bar{p}_T )\right\Vert_2
\]
where $\mathcal{X}$ is the object point cloud (for simplicity we will omit this from notation), $p_R$ and $p_T$ refer to the pose rotation and translation respectively.  
In other words, for a specific object, $\texttt{ADD}(p,\bar{p})$ is defined by the average 3D Euclidean distance between this object's two point clouds, corresponding to poses $p$ and $\bar{p}$.

\subsubsection{Uncertainty quantification and correlation analysis} 
For a pose estimator $g$, performance may vary with respect to different input images $x$.
Intuitively, an ideal uncertainty measurement would yield a high confidence (low uncertainty) for accurate pose predictions, and a low confidence (high uncertainty) for inaccurate predictions, such that we can trust $g$ in the scenarios when it performs well. For this reason, we aim to design a measurement $d$ that has a high correlation with the performance metric $\texttt{ADD}(p,\bar{p})$. The value of $d$ is treated as a UQ. 
This task is inherently challenging because the ground truth pose $\bar{p}$ is unknown and $x$ is in the \emph{target domain}, which may have a large gap to the training data (see \Cref{fig:ensemble} for examples).
 
To understand how well a measurement $d$ characterizes the true pose estimation error $\texttt{ADD}(p,\bar{p})$, we analyze the correlation between $d$ and the underlying pose estimation errors. The goal is to have a \emph{strong positive correlation} between them. In particular, suppose that there are $N$ images $\{x^{(1)},\cdots,x^{(N)}\}$ with pose predictions $\{p^{(1)}=g(x^{(1)}),\cdots,p^{(N)}=g(x^{(N)})\}$ and their ground truth poses $\{\bar{p}^{(1)},\cdots,\bar{p}^{(N)}\}$. We also compute the ADD errors of these predictions $\{\texttt{ADD}(p^{(1)},\bar{p}^{(1)}),\cdots,\texttt{ADD}(p^{(N)},\bar{p}^{(N)})\}$ and the uncertainty quantification $\{d^{(1)},\cdots,d^{(N)}\}$. 
We use the Spearman's rank correlation coefficient between $\{\texttt{ADD}(p^{(1)},\bar{p}^{(1)}),\cdots,\texttt{ADD}(p^{(N)},\bar{p}^{(N)})\}$ and $\{d^{(1)},\cdots,d^{(N)}\}$ as our metric to evaluate $d$. The Spearman's rank correlation coefficient is a nonparametric measure of rank correlation, {\em i.e.}, statistical dependence between the rankings of two variables. It assesses how well the relationship between two variables can be described using a monotonic function. Compared to the classic Pearson correlation, the Spearman's rank correlation supports any nonlinear dependence and it is also more robust~\cite{hauke2011comparison}.

In summary, the Spearman's rank metric is a scalar from $-1$ to $1$, and reflects the dependence of the pose estimation error $\texttt{ADD}(p,\bar{p})$ on the UQ $d$. In particular, a larger correlation coefficient suggests $d$ is a better uncertainty measurement, in the sense that a larger $d$ corresponds to a larger pose estimation error with higher chance.

\subsection{Heterogeneous Ensemble for Uncertainty Quantification}
\label{sec:ensemble}
Our ensemble method simultaneously infers the same image $x$ using $K$ different pose estimation models $g_1,\ldots,g_K$, as shown in \Cref{fig:ensemble}(a). 
Our method makes no assumption to these $K$ pose estimators. They are potentially heterogeneous in terms of model architecture and training data (see \Cref{sec:dope}). These $K$ models have $K$ different pose estimations: $\{p_1=g_1(x),\ldots,p_K=g_K(x)\}$. With these $K$ poses, we propose the following average disagreement based on the metric $f$:
\begin{equation}
d_f = \frac{1}{{K\choose 2}}\sum_{1 \leq i < j \leq K} f(p_i,p_j)
\label{eq:disagreement}
\end{equation}
where $f(p_i,p_j)$ is some distance metric between two poses $p_i$ and $p_j$, \textit{e.g.}, we can define $f(p_i,p_j)=\texttt{ADD}(p_i,p_j)$. 

In other words, $d_f$ represents the average disagreement of these $K$ different models against each other on a certain image input $x$. We use $d_f$ as an uncertainty quantification of these $K$ models on this image. The higher $d_f$ is, the more uncertain these $K$ models are about this image. For example, as shown in \Cref{fig:ensemble}(a), we use three models ($\modela$, $\modelb$, and $\modelc$) to detect the Milk object, and get three different poses represented by three bounding boxes with different colors. In the first example (left) in \Cref{fig:ensemble}(a), 
these three models disagree with each other much more than the second example (right) in \Cref{fig:ensemble}(a), which implies these three models are much more uncertain about the first example. 

The disagreement metric $f$ plays an important role in $d_f$. We examine two properties of different choices of $f$: \emph{object-variance} and \emph{model-variance}. 
\emph{Object-variance} allows flexibility across different objects, \textit{e.g.}, Milk and Ketchup may use different disagreement metrics.
\emph{Model-variance} allows flexibility across different pose estimators, \textit{e.g.}, $\modela$ and $\modelc$ may use different disagreement metrics.
We consider four types of disagreement metrics $f$ with different properties: 

\begin{itemize}
    \item \textbf{Translational disagreement} computes the translation error between $p_i$ and $p_j$ and ignores the rotation. It is an object-invariant and model-invariant disagreement, as the translation between $p_i$ and $p_j$ does not depend on object and model types.
    \item \textbf{Rotational disagreement} computes the rotation error between $p_i$ and $p_j$ and ignores the translation. It is also an object-invariant and model-invariant disagreement.
    \item \textbf{ADD disagreement} directly uses the $\texttt{ADD}(p_i,p_j)$ metric as $f(p_i,p_j)$. It is object-variant but model-invariant, as computing the ADD metric requires the object 3D geometry, but does not depend on the pose estimator. Empirically, we find that ADD disagreement is much better than translational/rotational disagreement.
    \item \textbf{Learned disagreement} is a data-driven disagreement. In this case, we need a dataset of labeled images in the target domain to train $f(p_i,p_j)$, which is represented by a neural network (see \Cref{sec:learned} for details). Note that we can train different $f$ for different pose estimators and objects, so this disagreement allows full flexibility across objects and models, \textit{i.e.}, it is object-variant and model-variant.
\end{itemize}


\subsection{Learned Disagreement Metric for Ensemble}
\label{sec:learned}
Differently from the translational/rotational/ADD disagreement metrics, the learned disagreement metric is data driven and needs labeled data in the target domain. The idea of learning to combine different models is inspired by model stacking in statistics~\cite{wolpert1992stacked}, where the goal is to boost the performance of an ensemble of heterogeneous models by learning a combination of their outputs in the target domain. 
In particular, for a certain object in the target domain we are given images $\{x^{(1)},\ldots,x^{(M)}\}$ with pose supervision $\{\bar{p}^{(1)},\ldots,\bar{p}^{(M)}\}$. With $K$ models/pose estimators $g_1,\ldots,g_K$ in the ensemble, we have their pose predictions on each image in $\{x^{(1)},\ldots,x^{(M)}\}$:
\begin{equation*}
    \left\{p^{(m)}_1=g_1(p^{(m)}),\ldots,p^{(m)}_K=g_K(p^{(m)})\right\}_{m=1}^M
\end{equation*}
For model $g_k$ we also compute its ADD errors:
\begin{equation*}
    \left\{ \texttt{ADD}(p_k^{(m)},\bar{p}^{(m)}) \right\}_{m=1}^M
\end{equation*}
With these $K$ models' predictions and the ADD errors of the model $g_k$, our goal is to minimize the following loss function with respect to the disagreement metric $f_k$:
\begin{equation}
    \begin{small}
    \label{eq:training}
    \begin{aligned}
        \min_{f_k\in\mathcal{F}} \sum_{m=1}^M \bigg| &\frac{\sum_{1\leq i < j\leq K} f_k(p_i^{(m)},p_j^{(m)})}{{K\choose 2}} -\texttt{ADD}(p_k^{(m)},\bar{p}^{(m)}) \bigg|^2
    \end{aligned}
    \end{small}
\end{equation}
where $\mathcal{F}$ is the function class.
\Cref{eq:training} finds the optimal disagreement metric $f^*_k$ for the pose estimator $g_k$, given $M$ labeled data in the target domain. 
Note that different pose estimators may have different optimal disagreement metrics, {\em i.e.}, $f^*_k$ depends on $k$ and the learned disagreement is model-variant (see the definition in \Cref{sec:ensemble}). 
In practice, we use a simple feed-forward neural network to represent $f_k$ and stochastic gradient descend is used to optimize the loss function in \Cref{eq:training}. 
Empirically, we find that training with $\{x^{(1)},\ldots,x^{(M)}\}$ learns a disagreement metric that can generalize to other images in the target domain, and
that it outperforms all other model-invariant and learning-free disagreements (translational/rotational/ADD).

\section{Implementation details}
\label{sec:dope}
For UQ, our ensemble method uses multiple pose estimators from possibly different 
biases, \textit{e.g.}, different training datasets. 
We use the newly available real-world HOPE dataset~\cite{tyree2019hope} (see \Cref{fig:ensemble}(a) for examples),
which consists of 28 grocery toy items with associated 3D models. 
It has 238 unique images with 914 unique object poses for evaluation, and
the dataset does not provide any training images.
For each object presented in the HOPE dataset, we generated different sets of 60,000 images, which 
uses visual domain randomization~\cite{tremblay2017icra:cube} to bridge the reality gap.
We generated two different datasets from two different synthetic data generators available online, NDDS~\cite{to2018ndds} (see \Cref{fig:ensemble}(c) for one example) and NViSII~\cite{Morrical20visii} (see \Cref{fig:ensemble}(b)). 
The former one uses rasterization whereas the latter one uses ray-tracing. 
%
We trained the open-source pose estimator DOPE~\cite{tremblay2018pose} on both datasets as well as a modified DOPE architecture on NDDS only. It is worth noting that our UQ method is compatible with any pose estimator architecture and arbitrary number of estimators. In this paper we use DOPE for its state-of-the-art RGB-based pose estimation performance.

The DOPE pose estimator first finds the projected cuboid keypoints in the image frame, and using the known size of the object and
the camera intrinsics it regresses to the object pose using P$n$P~\cite{gao:pnp}. 
The keypoints are represented as heatmaps, more precisely, nine heapmaps for the cuboid vertex and its centroid.
A greedy algorithm finds the keypoint positions in the original image. 
For each model we report the area under the curve (AUC) for ADD threshold curves at 10 cm for all objects in the HOPE dataset.   
In this work we use the following models:
\begin{itemize}
    \item \textbf{$\modela$ model.} 
    We directly use the model architecture in~\cite{tremblay2018pose} and train all 28 models on NDDS synthetic data with domain randomization.
    This model achieves an AUC of 0.37.
    \item \textbf{$\modelb$ model.} Inspired by previous work~\cite{lee2020icra:dream}, we use the auto-encoder architecture that has more capacity than the original DOPE model architecture. This model achieves an AUC of 0.45.
    \item \textbf{$\modelc$ model.} This model has the same architecture to the $\modela$ model, but is trained on NViSII synthetic data.
    This model achieves an AUC of 0.37.
\end{itemize}

We compare the ensemble method against two baselines. Both baselines are based on the internal UQ of a single neural network model, \textit{i.e.}, these two baselines do not leverage information from multiple models. The baselines include:
\begin{itemize}
    \item \textbf{Neural network confidence score (Confidence).} Along with the predicted 6-DoF pose, the DOPE model~\cite{tremblay2018pose} also gives a confidence score to each pose estimation. In particular, this confidence score reflects the confidence level of the centroid of the detected object, {\em e.g.}, the height of the peak on the centroid's heatmap. 
    \item \textbf{Randomly sampling belief maps (GUAPO).} This baseline~\cite{lee2020guided} augments the peak estimation algorithm in DOPE by fitting a 2D Gaussian around each found peak. GUAPO then runs P$n$P algorithm on $T$ sets of keypoints, where each set of keypoint is constructed by sampling from all the 2D Gaussians. This provides $T$ possible poses of the object consistent with the detection algorithm, and the variance of these $T$ possible poses (standard deviation) represents the uncertainty. 
\end{itemize}


%% file: experiments.tex
\section{Experiments}
\begin{figure}[t]
    \centering
    \vspace{2mm}
    \includegraphics[width=1.0\linewidth]{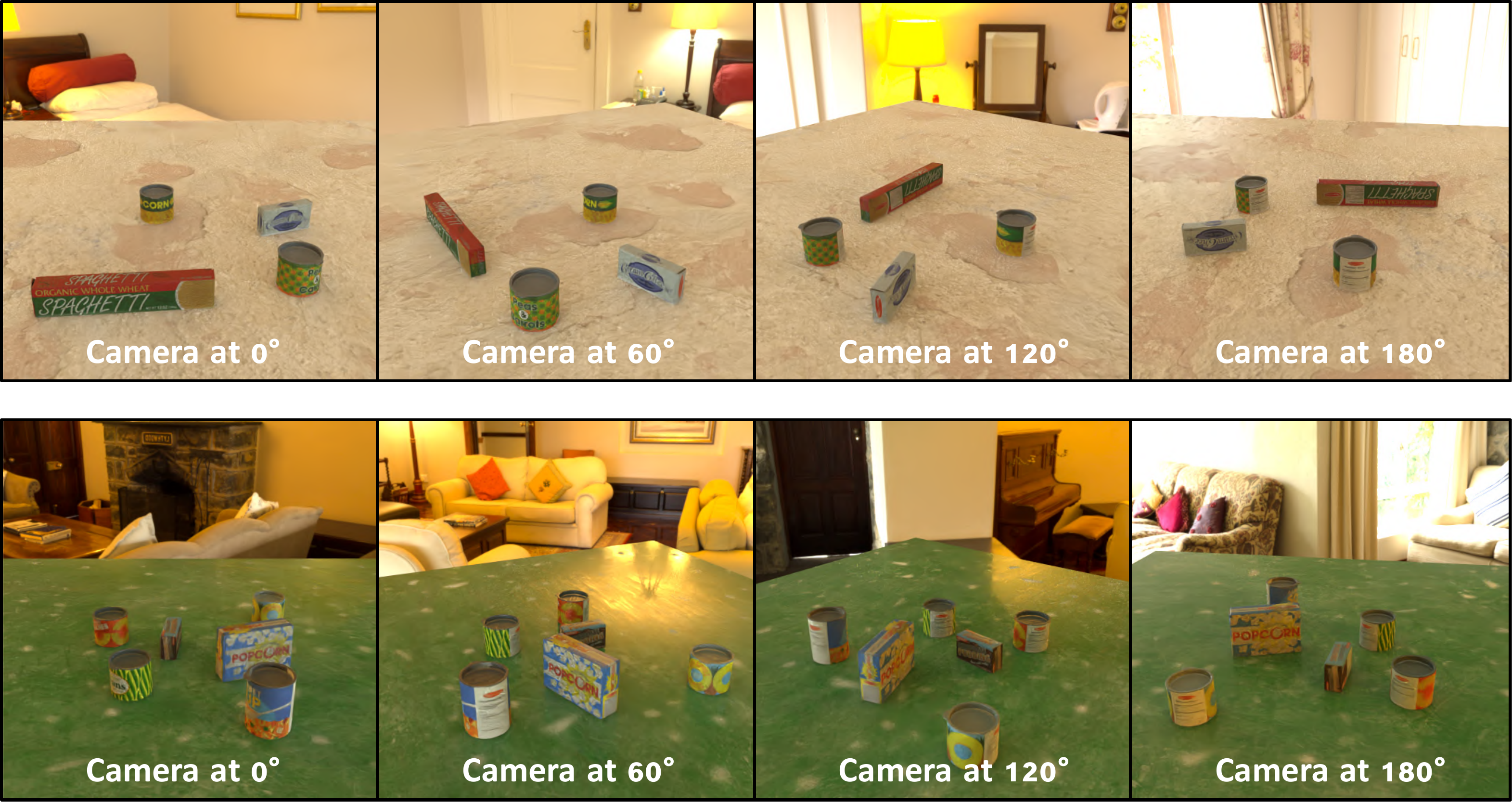}
    \vspace{-22pt}
    \caption{Two test sequences generated by NViSII~\cite{Morrical20visii}. Each sequence contains 45 images with camera angle from $0^{\circ}$ to $180^{\circ}$.}
    \label{fig:visii_example}
    \vspace{-12pt}
\end{figure}

We evaluate the capabilities of our uncertainty quantification (UQ) approach with respect to the problem of 6-DoF pose estimation of known objects. 
In other words, we want to evaluate how likely a 3D position and orientation prediction is correct, and to that end, we seek to answer the following questions throughout the experiments:
1) How does our method compare to different ways to quantify uncertainty?  
2) Can our method be used to ensure higher success-rate for grasping objects with a real-world manipulator?


\subsection{Correlation Analysis}

We evaluate our proposed method using the correlation between UQ values and the predicted pose errors against ground truth (see the Spearman's rank correlation coefficient in \Cref{sec:spearman}).
This evaluation is conducted on the real-world HOPE dataset~\cite{tyree2019hope}. Please note that all our pose estimation models are only trained on synthetic data (either NDDS or NViSII). In particular, for each object in the 28 HOPE set, we collected all images in the HOPE dataset~\cite{tyree2019hope} that do not contain multiple instances of the same object. 
We first explore which disagreement metric $f$ in \cref{eq:disagreement} offers the best correlation, and we explore the impact of different 
architectures/training data when using ensemble for UQ.

\subsubsection{Disagreement metric selection}
We evaluate six UQ methods: our ensemble-based method with four types of disagreement (ADD, rotation, translation, learned) as described in \Cref{sec:ensemble}, and the two baselines presented in \Cref{sec:dope}. 
For the ensemble methods we use two heterogeneous models with different architectures ($\modela$ and $\modelb$) in the ensemble.

For the learned disagreement, we train two neural networks on $1/3$ of the HOPE dataset for each object to learn the disagreement metrics $f$ for both $\modela$ and $\modelb$ (see \Cref{sec:learned}). 
We use four-layer multilayer perceptrons with ReLU activations to parameterize the disagreement metric, where the input dimension is 14 (two poses represented by quaternions and 3D vectors) and the output is a scalar (disagreement between two poses). 
The other five methods do not need labeled data.


Table~\ref{tab:ensemble_result} shows the correlation coefficients for the six UQ methods computed using $\modela$ and $\modelb$.
From this experiment we see that 1) ensemble methods with all four disagreements perform better than baselines, except for an outlier from the rotational disagreement; 2) the learned disagreement achieves the best correlations because it is trained on the target domain and it is both object-variant and model-variant, \emph{i.e.}, different objects and models may learn diverse disagreements accordingly; and 3) ensemble with ADD disagreement yields the best results in all data-free methods, and is only slightly worse than the learned disagreement ensemble method. 
The small performance gap between the \emph{data-free} ADD disagreement and learned disagreement suggests that ADD is a suitable disagreement for UQ in an ensemble.

\subsubsection{Architecture and training data}
The second set of experiments considers the influence of heterogeneity from training data/architectures, and \Cref{tab:ensemble_result_visii} shows the results. Note that the $\modelc$ model is trained on the NViSII synthetic data, but the $\modela$ and $\modelb$ models are trained on the NDDS synthetic data. Here we focus on the performance of the ADD disagreement ensemble with three different ensemble combinations: ($\modelc$, $\modela$), ($\modelc$, $\modelb$), and ($\modelc$, $\modela$, $\modelb$). More precisely, we evaluate two baselines and ADD disagreement ensemble with these three combinations, as their correlation with respect to the $\modelc$ model.

\Cref{tab:ensemble_result_visii} shows the results on this experiment.
We find that 1) all the ensemble combinations have much stronger correlations than the baselines; 2) the three-model ensemble is slightly worse than two-model ensembles, which implies adding more models in the ensemble does not necessarily improves UQ; and 3) compared with \Cref{tab:ensemble_result}, $\modelc$ achieves better correlations than $\modela$ and $\modelb$. 
In particular, the baseline method GUAPO for $\modelc$ already outperforms ensemble methods for $\modela$ or $\modelb$ in \Cref{tab:ensemble_result}, which implies it is easier to quantify uncertainty for the NViSII~\cite{Morrical20visii} (based on ray-tracing) trained model than the NDDS~\cite{tremblay2017icra:cube} (based on rasterization) trained models, possibly due to realistic light modelling from ray-tracing~\cite{Morrical20visii}.

\begin{table}[t]
\begin{center}
\vspace{2mm}
\caption{Correlation analysis for different disagreements}
\begin{tabular}{lcc}
& $\modela$ & $\modelb$ \\
\hline
Confidence~\cite{tremblay2018pose} & 0.26 $\pm$ 0.14 & 0.40 $\pm$ 0.16 \\
GUAPO~\cite{lee2020guided} & 0.45 $\pm$ 0.18 & 0.44 $\pm$ 0.15 \\
\hdashline
Translational Disagreement & 0.52 $\pm$ 0.11 & 0.46 $\pm$ 0.12 \\
Rotational Disagreement & 0.39 $\pm$ 0.15 & 0.44 $\pm$ 0.13 \\
ADD Disagreement & \textbf{0.55} $\pm$ 0.11 & \textbf{0.50} $\pm$ 0.12 \\
Learned Disagreement$^{\text{*}}$ & 0.58 $\pm$ 0.09 & 0.52 $\pm$ 0.13 \\
\label{tab:ensemble_result}
\end{tabular}
\end{center}
\vspace{-4mm}
\hspace{2mm}
\begin{scriptsize}
$^{\text{*}}$requires labeled data in the target domain
\end{scriptsize}
\vspace{-1mm}
\end{table}

\begin{table}[t]
\centering
\caption{Correlation analysis for architectures and training data}
\begin{tabular}{l|c}
& $\modelc$ \\
\hline
Confidence~\cite{tremblay2018pose} & 0.38 $\pm$ 0.16 \\
GUAPO~\cite{lee2020guided} & 0.55 $\pm$ 0.15 \\
\hdashline
ADD ($\modelc$, $\modela$) & 0.64 $\pm$ 0.14 \\
ADD ($\modelc$, $\modelb$) & \textbf{0.68} $\pm$ 0.14 \\
ADD (All three models) & 0.62 $\pm$ 0.14 \\
\end{tabular}
\label{tab:ensemble_result_visii}
\vspace{\shrink pt}
\end{table}


\begin{figure*}[t]
  \centering
  \vspace{2mm}
  \includegraphics[width = 1.0\textwidth]{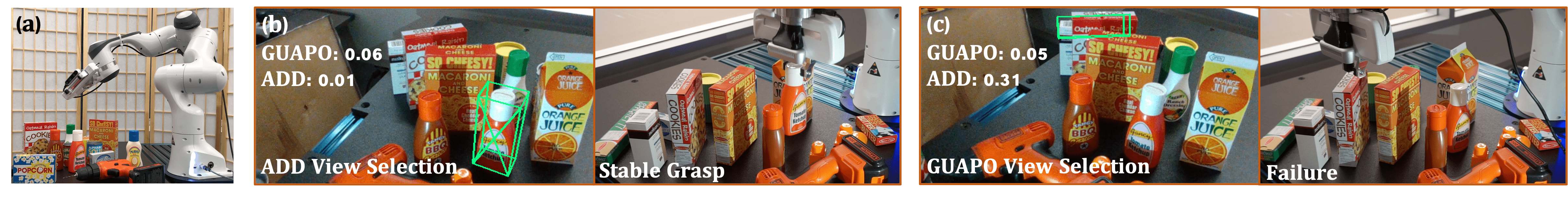}
  \vspace{-17pt}
  \caption{(a) Experimental setup. (b-c) The view selection and grasping results of the Ketchup object from our method and GUAPO~\cite{lee2020guided}. We use the $\modela$ model (green box) as the pose estimator and we quantify its uncertainty. For our method we use the average ADD disagreement between $\modela$ and $\modelb$ as the uncertainty. The numbers in (b-c) show the uncertainty quantification values of GUAPO and our method.}
  \label{fig:grasp}
  \vspace*{-2mm}
\end{figure*}

\subsection{Application I: Camera Perspective Selection}
The estimate uncertainty can inform an active perception agent to explore an environment and select the most confident observation for decision making. In the context of pose estimation, visual observations from different camera perspectives may result in object occlusions and other environmental noises. 
Here we examine whether our UQ can select an optimal perspective for making the best pose predictions.

To quantitatively study the performances of different UQ methods in this problem, we generate 125 image sequences using NViSII~\cite{Morrical20visii} (see \Cref{fig:visii_example} for examples). In particular, for each sequence, we randomly pick 3--6 objects from 28 HOPE objects, and randomize their collision-free placements. We also randomize the background, table texture, and illumination for each sequence. Finally, we move around the virtual camera in 180 degree and sample 45 images. 

\begin{table}[t]
\centering
\caption{Average ADD error of the most confident frame (unit: cm)}
\vspace{-2.5mm}
\begin{tabular}{l|cc}
& $\modela$ & $\modelb$ \\
\hline
Confidence~\cite{tremblay2018pose} & 4.9 $\pm$ 6.5 & 4.7 $\pm$ 6.1 \\
GUAPO~\cite{lee2020guided} & 5.6 $\pm$ 6.9 & 5.1 $\pm$ 5.9 \\
ADD (Ours) & \textbf{3.5} $\pm$ \textbf{4.2} & \textbf{3.1} $\pm$ \textbf{3.6} \\
\hdashline
Oracle & 1.0 $\pm$ 1.2 & 1.2 $\pm$ 1.4 \\
\end{tabular}
\label{tab:visii_result}
\vspace{-14pt}
\end{table}

We use a simple greedy approach to select the optimal frame in the image sequence. For each object in each sequence, we directly select the frame of the least UQ value. For instance, in the case of the ensemble method, we will choose the object with the smallest $d_f$ in \cref{eq:disagreement}. Other than the ensemble method and two baselines, we also compute the performance of the oracle approach, which has access to the ground truth and will directly select the image with lowest ADD error. This oracle helps us understand the upper bound of the performance of the pose estimators. 
We use ADD as the disagreement metric in ensemble and we excluded $\modelc$ model from this experiment since it is trained on images from NViSII renderer. 

The results are presented in \Cref{tab:visii_result}, where we average all 28 objects and all 125 sequences. 
Our method decreases the ADD error of the greedy choice by 30--40\% compared to baselines, and the variance is also significantly reduced.

\subsection{Application II: Real-World Robotic Grasping}

To examine the utility of UQ of our proposed ensemble method for downstream tasks in the real world, we use an uncertainty-guided robotic grasping task to demonstrate its improvement over the baselines. 
We use a 7-DoF robotic arm, Franka Emika Panda, for our grasping task, see \Cref{fig:grasp}(a). 
Similar to the experiments on NViSII synthetic data, we first control the robot to observe 6 images of the workspace with a clutter of objects from different perspectives, and we use the UQ value to select the most confident point of view. The pose estimation and UQ process of all 6 images in a batch takes less than one second on an NVIDIA RTX 2080 GPU. We command the robot to grasp the target object using the pose estimation from this point of view, see \Cref{fig:grasp}(b-c). After robot execution, we evaluate if the grasp is a stable grasp, unstable grasp, or a complete failure. Unstable grasp corresponds to the cases where the gripper successfully makes contact with the object, but it slipped away. For failure cases, the gripper fails to touch the target object at all. 
We use the $\modela$ model as our pose estimator, and we use the average ADD disagreement between $\modela$ and $\modelb$ for UQ.

\begin{table}[t]
\centering

\caption{Uncertainty-guided real-world robot grasping success rate}
\vspace{-2.5mm}
\begin{tabular}{l|ccc}
& Confidence~\cite{tremblay2018pose} & GUAPO~\cite{lee2020guided} & ADD (Ours)\\
\hline
Stable Grasp & 35\% & 25\% & \textbf{90\%} \\
Unstable Grasp & 10\% & 25\% & \textbf{10\%} \\
Failure & 55\% & 50\% & \textbf{0\%} \\
\end{tabular}
\label{tab:robotic_grasping}
\vspace{-14pt}
\end{table}


We evaluate on four objects (Ketchup, BBQ Sauce, Mayo, and Butter) of diverse shapes/colors/textures, and attempt 5 trials for each object and each method. For a fair comparison, we evaluate all three methods in the same environment setting for each trial. 
A summary of results is shown in Table~\ref{tab:robotic_grasping}, 
suggesting that our method significantly outperforms the two baselines. \Cref{fig:grasp}(b) shows a Ketchup example of view selection and grasping from our method, which leads to a successful grasping. \Cref{fig:grasp}(c) shows the result from GUAPO under the same environment, where GUAPO is overconfident about a non-Ketchup pose estimation.

%% file: conclusions.tex
\section{Conclusions}
We develop a simple, efficient, and plug-and-play method for quantifying the predictive uncertainty of 6-DoF pose estimators. Our method uses an ensemble of two or more heterogeneous models and computes their average disagreement against one another as the uncertainty quantification. Our experimental results demonstrate that average distance (ADD) is a suitable learning-free disagreement metric, and the proposed method significantly outperforms baselines in different tasks, including real-robot grasping. Future directions include the theoretical analysis of ensemble-based UQ and the integration of our method with other algorithms, such as uncertainty-aware RL and pose estimation in videos.


\section*{Acknowledgement}
We would like to thank members of the NVIDIA AI Algorithms research team for their constructive feedback and Nathan Morrical for his help with the NViSII renderer.


%% file: bib.bib
@article{Andrieu2003,
  year={2003},
  title = "An Introduction to {MCMC} for Machine Learning",
  author = "Andrieu, Christophe and de Freitas, Nando and Doucet, Arnaud and Jordan, Michael I.",
  year = "2003",
  issn = "0885-6125",
  journal = "Machine Learning",
  number = "1-2",
  pages = "5-43",
  publisher = "Kluwer Academic Publishers",
  volume = "50",
}

@inproceedings{cao2017cvpr:mppaf,
  author    = {Z. Cao and T. Simon and S.-E. Wei and Y. Sheikh},
  title     = {Realtime Multi-Person 2{D} Pose Estimation using Part Affinity Fields},
  booktitle = {CVPR},
  year      = 2017
}

@inproceedings{dhall2019arx:cones,
  title     = {Real-time {3D} Traffic Cone Detection for Autonomous Driving},
  author    = {Ankit Dhall and Dengxin Dai and Luc Van Gool},
  booktitle = {arXiv:1902.02394},
  year      = 2019
}

@article{gao:pnp,
  author  = {Xiao-Shan Gao and Xiao-Rong Hou and Jianliang Tang and Hang-Fei Cheng},
  journal = {IEEE Transactions on Pattern Analysis and Machine Intelligence},
  title   = {Complete solution classification for the perspective-three-point problem},
  year    = {2003},
  volume  = {25},
  number  = {8},
  pages   = {930--943},
  month   = aug
}

@inproceedings{hinterstoisser2012model,
  title     = {Model based training, detection and pose estimation of texture-less {3D} objects in heavily cluttered scenes},
  author    = {Hinterstoisser, Stefan and Lepetit, Vincent and Ilic, Slobodan and Holzer, Stefan and Bradski, Gary and Konolige, Kurt and Navab, Nassir},
  booktitle = {ACCV},
  year      = 2012
}

@inproceedings{hodan2017wacv:tless,
  title     = {{T-LESS}: An {RGB-D} Dataset for {6D} Pose Estimation of Texture-less Objects},
  author    = {Hoda{\v{n}}, T. and Haluza, P. and Obdr{\v{z}}{\'a}lek, {\v{S}}. and Matas, J. and Lourakis, M. and Zabulis, X.},
  booktitle = {WACV},
  year      = 2017
}

@inproceedings{hu2019segmentation,
  title     = {Segmentation-driven {6D} object pose estimation},
  author    = {Hu, Yinlin and Hugonot, Joachim and Fua, Pascal and Salzmann, Mathieu},
  booktitle = {IEEE Conference on Computer Vision and Pattern Recognition},
  pages     = {3385--3394},
  year      = {2019}
}

@article{Jordan99,
  author = {Jordan, Michael I. and Ghahramani, Zoubin and Jaakkola, Tommi S. and Saul, Lawrence K.},
  journal = {Machine Learning},
  title = {An Introduction to Variational Methods for Graphical Models.},
  volume = 37,
  year = 1999
}

@inproceedings{kim2019icra:insertion,
  title     = {Shallow-Depth Insertion: Peg in Shallow Hole through Robotic In-Hand Manipulation},
  author    = {Chung Hee Kim and Jungwon Seo},
  booktitle = {ICRA},
  year      = 2019
}

@article{lepetit2009ijcv:epnp,
  author  = {V. Lepetit and F. Moreno-Noguer and P. Fua},
  title   = {{EP\emph{n}P}: {A}n Accurate {O}(\emph{n}) Solution to the {P\emph{n}P} Problem},
  journal = {International Journal Computer Vision},
  volume  = {81},
  number  = {2},
  year    = {2009}
}

@article{li2019rethinking,
  title   = {Rethinking on Multi-Stage Networks for Human Pose Estimation},
  author  = {Li, Wenbo and Wang, Zhicheng and Yin, Binyi and Peng, Qixiang and Du, Yuming and Xiao, Tianzi and Yu, Gang and Lu, Hongtao and Wei, Yichen and Sun, Jian},
  journal = {arXiv preprint arXiv:1901.00148},
  year    = {2019}
}

@inproceedings{liu2018:robotsafe,
  title     = {Robot safe interaction system for intelligent industrial co-robots},
  author    = {Liu, Changliu and Tomizuka, Masayoshi},
  booktitle = {arXiv:1808.03983},
  year      = 2018
}

@inproceedings{peng2019pvnet,
  title     = {PVNet: Pixel-wise Voting Network for 6DoF Pose Estimation},
  author    = {Peng, Sida and Liu, Yuan and Huang, Qixing and Zhou, Xiaowei and Bao, Hujun},
  booktitle = {CVPR},
  year      = {2019}
}

@inproceedings{sun2019deep,
  title     = {Deep High-Resolution Representation Learning for Human Pose Estimation},
  author    = {Sun, Ke and Xiao, Bin and Liu, Dong and Wang, Jingdong},
  booktitle = {CVPR},
  year      = {2019}
}

@inproceedings{Sundermeyer_2018_ECCV,
  author    = {Sundermeyer, Martin and Marton, Zoltan-Csaba and Durner, Maximilian and Brucker, Manuel and Triebel, Rudolph},
  title     = {Implicit 3D Orientation Learning for 6D Object Detection from RGB Images},
  booktitle = {ECCV},
  year      = {2018}
}

@article{tan20176d,
  title={6d object pose estimation with depth images: A seamless approach for robotic interaction and augmented reality},
  author={Tan, David Joseph and Navab, Nassir and Tombari, Federico},
  journal={arXiv preprint arXiv:1709.01459},
  year={2017}
}

@inproceedings{tekin2018cvpr:objpose,
  author    = {Bugra Tekin and Sudipta N. Sinha and Pascal Fua},
  title     = {Real-Time Seamless Single Shot {6D} Object Pose Prediction},
  booktitle = {CVPR},
  year      = 2018
}

@inproceedings{tian2019icra:fog,
  title     = {A Fog Robotic System for Dynamic Visual Servoing},
  author    = {Nan Tian and Ajay Kummar Tanwani and Jinfa Chen and Mas Ma and Robert Zhang and Bill Huang
Ken Goldberg and Somayeh Sojoudi},
  booktitle = {ICRA},
  year      = 2019
}

@misc{to2018ndds,
  author = {Thang To and Jonathan Tremblay and Duncan McKay and Yukie Yamaguchi and Kirby Leung and Adrian Balanon and Jia Cheng and Stan Birchfield},
  note   = { \url{https://github.com/NVIDIA/Dataset\_Synthesizer} },
  title  = {{NDDS}: {NVIDIA} Deep Learning Dataset Synthesizer},
  year   = {2018}
}

@inproceedings{tremblay2017icra:cube,
  author    = {Jonathan Tremblay and Thang To and Artem Molchanov and Stephen Tyree and 
 Jan Kautz and Stan Birchfield},
  title     = {Synthetically Trained Neural Networks for Learning Human-Readable Plans from Real-World Demonstrations},
  booktitle = {ICRA},
  year      = 2018
}

@inproceedings{tremblay2018pose,
  title     = {Deep Object Pose Estimation for Semantic Robotic Grasping of Household Objects},
  author    = {Jonathan Tremblay and Thang To and Balakumar Sundaralingam and Yu Xiang and Dieter Fox and Stan Birchfield},
  booktitle = {CoRL},
  year      = {2018}
}

@inproceedings{wei2016cvpr:cpm,
  author    = {S.-E. Wei and V. Ramakrishna and T. Kanade and Y. Sheikh},
  title     = {Convolutional Pose Machines},
  booktitle = {CVPR},
  year      = 2016
}

@inproceedings{xiang2018rss:posecnn,
  author    = {Y. Xiang and T. Schmidt and V. Narayanan and D. Fox},
  title     = {Pose{CNN}: {A} Convolutional Neural Network for {6D} Object Pose Estimation in Cluttered Scenes},
  booktitle = {RSS},
  year      = 2018
}

@inproceedings{xiao2018simple,
  title     = {Simple baselines for human pose estimation and tracking},
  author    = {Xiao, Bin and Wu, Haiping and Wei, Yichen},
  booktitle = {European Conference on Computer Vision (ECCV)},
  pages     = {466--481},
  year      = {2018}
}

@article{zakharov2019dpod,
  title   = {{DPOD}: {D}ense {6D} pose object detector in {RGB} images},
  author  = {Zakharov, Sergey and Shugurov, Ivan and Ilic, Slobodan},
  journal = {arXiv preprint arXiv:1902.11020},
  year    = {2019}
}

@article{deng2019poserbpf,
  title={{PoseRBPF: A Rao-Blackwellized particle filter for 6d object pose tracking}},
  author={Deng, Xinke and Mousavian, Arsalan and Xiang, Yu and Xia, Fei and Bretl, Timothy and Fox, Dieter},
  journal={arXiv preprint arXiv:1905.09304},
  year={2019}
}

@inproceedings{lee2020icra:dream,
  title={Camera-to-Robot Pose Estimation from a Single Image},
  author={Lee, Timothy E and Tremblay, Jonathan and To, Thang and Cheng, Jia and Mosier, Terry and Kroemer, Oliver and Fox, Dieter and Birchfield, Stan},
  booktitle={International Conference on Robotics and Automation},
  year=2020,
}

@article{yang2020human,
  title={Human Grasp Classification for Reactive Human-to-Robot Handovers},
  author={Yang, Wei and Paxton, Chris and Cakmak, Maya and Fox, Dieter},
  journal={arXiv preprint arXiv:2003.06000},
  year={2020}
}

@inproceedings{handa2020dexpilot,
  title={{DexPilot}: Vision-Based Teleoperation of Dexterous Robotic Hand-Arm System},
  author={Handa, Ankur and Van Wyk, Karl and Yang, Wei and Liang, Jacky and Chao, Yu-Wei and Wan, Qian and Birchfield, Stan and Ratliff, Nathan and Fox, Dieter},
  booktitle={IEEE International Conference on Robotics and Automation},
  year={2020},
}

@incollection{Lakshminarayanan:ensemble,
title = {Simple and Scalable Predictive Uncertainty Estimation using Deep Ensembles},
author = {Lakshminarayanan, Balaji and Pritzel, Alexander and Blundell, Charles},
booktitle = {Advances in Neural Information Processing Systems},
editor = {I. Guyon and U. V. Luxburg and S. Bengio and H. Wallach and R. Fergus and S. Vishwanathan and R. Garnett},
year = {2017},
}

@article{amodei2016concrete,
  title={Concrete problems in {AI} safety},
  author={Amodei, Dario and Olah, Chris and Steinhardt, Jacob and Christiano, Paul and Schulman, John and Man{\'e}, Dan},
  journal={arXiv preprint arXiv:1606.06565},
  year={2016}
}

@article{kalmanfilter,
    Author = {Kalman, Rudolph Emil},
    Title = {A New Approach to Linear Filtering and Prediction Problems},
    Journal = {Transactions of the ASME--Journal of Basic Engineering},
    Volume = {82},
    Pages = {35--45},
    Year = {1960}
}

@article{blum2019fishyscapes,
  title={The fishyscapes benchmark: Measuring blind spots in semantic segmentation},
  author={Blum, Hermann and Sarlin, Paul-Edouard and Nieto, Juan and Siegwart, Roland and Cadena, Cesar},
  journal={arXiv preprint arXiv:1904.03215},
  year={2019}
}

@book{bernardo2009bayesian,
  title={Bayesian theory},
  author={Bernardo, Jos{\'e} M and Smith, Adrian FM},
  volume={405},
  year={2009},
  publisher={John Wiley \& Sons}
}

@misc{Morrical20visii,
author = {Nathan Morrical and Jonathan Tremblay and Stan Birchfield and Ingo Wald},
note= {\url{ https://github.com/owl-project/NVISII/ }},
title = {{NVISII}: NVIDIA Scene Imaging Interface},
Year = 2020
}

@article{lee2020guided,
  title={Guided Uncertainty-Aware Policy Optimization: Combining Learning and Model-Based Strategies for Sample-Efficient Policy Learning},
  author={Lee, Michelle A and Florensa, Carlos and Tremblay, Jonathan and Ratliff, Nathan and Garg, Animesh and Ramos, Fabio and Fox, Dieter},
  journal={arXiv preprint arXiv:2005.10872},
  year={2020}
}

@inproceedings{tyree2019hope,
  title={6-{DoF} pose estimation of household objects for robotic manipulation: an accessible dataset and benchmark},
  author={Stephen Tyree and Jonathan Tremblay and Thang To and Jia Cheng and Terry Mossier and Stan Birchfield},
  booktitle={ICCV Workshop on Recovering 6D Object Pose},
  year={2019}
}

@article{hauke2011comparison,
  title={Comparison of values of Pearson's and Spearman's correlation coefficients on the same sets of data},
  author={Hauke, Jan and Kossowski, Tomasz},
  journal={Quaestiones geographicae},
  volume={30},
  number={2},
  pages={87--93},
  year={2011},
  publisher={Sciendo}
}

@inproceedings{wang2019densefusion,
  title={{DenseFusion}: 6d object pose estimation by iterative dense fusion},
  author={Wang, Chen and Xu, Danfei and Zhu, Yuke and Martin-Martin, Roberto and Lu, Cewu and Fei-Fei, Li and Savarese, Silvio},
  booktitle={IEEE Conference on Computer Vision and Pattern Recognition},
  year={2019}
}

@article{loquercio2020general,
  title={A general framework for uncertainty estimation in deep learning},
  author={Loquercio, Antonio and Segu, Mattia and Scaramuzza, Davide},
  journal={IEEE Robotics and Automation Letters},
  volume={5},
  number={2},
  pages={3153--3160},
  year={2020},
  publisher={IEEE}
}

@inproceedings{doersch2019sim2real,
  title={Sim2real transfer learning for 3D human pose estimation: motion to the rescue},
  author={Doersch, Carl and Zisserman, Andrew},
  booktitle={Advances in Neural Information Processing Systems},
  year={2019}
}

@article{lu2020robust,
  title={Robust Keypoint Detection and Pose Estimation of Robot Manipulators with Self-Occlusions via Sim-to-Real Transfer},
  author={Lu, Jingpei and Richter, Florian and Yip, Michael},
  journal={arXiv preprint arXiv:2010.08054},
  year={2020}
}

@inproceedings{gal2016dropout,
  title={Dropout as a bayesian approximation: Representing model uncertainty in deep learning},
  author={Gal, Yarin and Ghahramani, Zoubin},
  booktitle={International Conference on Machine Learning},
  year={2016}
}

@inproceedings{zeng2017multi,
  title={Multi-view self-supervised deep learning for 6d pose estimation in the {Amazon Picking Challenge}},
  author={Zeng, Andy and Yu, Kuan-Ting and Song, Shuran and Suo, Daniel and Walker, Ed and Rodriguez, Alberto and Xiao, Jianxiong},
  booktitle={IEEE International Conference on Robotics and Automation},
  year={2017},
}

@article{wolpert1992stacked,
  title={Stacked generalization},
  author={Wolpert, David H},
  journal={Neural networks},
  volume={5},
  number={2},
  pages={241--259},
  year={1992},
  publisher={Elsevier}
}

@article{chen2019grip,
  title={Grip: Generative robust inference and perception for semantic robot manipulation in adversarial environments},
  author={Chen, Xiaotong and Chen, Rui and Sui, Zhiqiang and Ye, Zhefan and Liu, Yanqi and Bahar, R and Jenkins, Odest Chadwicke},
  journal={IROS},
  year={2019}
}
